\documentclass[runningheads]{llncs}
\usepackage[title]{appendix}
\usepackage{graphicx}
\usepackage[utf8]{inputenc} 
\usepackage[T1]{fontenc}    
\usepackage{hyperref}       

\usepackage{booktabs}       
\usepackage{amsfonts}       
\usepackage{nicefrac}       
\usepackage{microtype}      

\usepackage{amsmath,amsfonts,cancel,subfigure,pifont,tikz,wrapfig}
\usetikzlibrary{arrows,shapes}
\usepackage{bbold}

\newcommand{\figref}[1]{Fig. \ref{#1}}
\newcommand{\secref}[1]{Sect. \ref{#1}}
\renewcommand{\eqref}[1]{Eq. (\ref{#1})}
\newcommand{\appref}[1]{Appendix \ref{#1}}
\newcommand\independent{\protect\mathpalette{\protect\independenT}{\perp}} 
\def\independenT#1#2{\mathrel{\rlap{$#1#2$}\mkern2mu{#1#2}}}
\newcommand\ci{\independent}
\newcommand{\av}[1]{\langle #1 \rangle}
\definecolor{myblue}{cmyk}{0,0.9,0.9,0.1}
\newcommand{\myblue}{\color{myblue}}

\tikzstyle{ocont}=[ellipse,draw=black!100,thick,minimum size=6mm,>=stealth]  
\tikzstyle{dgraph}=[->, line width=1.5pt]

\newcommand{\ie}{\emph{i.e.}}
\newcommand{\eg}{\emph{e.g.}}
\newcommand{\CBN}{CBN}
\newcommand{\CBNs}{CBNs}
\newcommand{\PSE}{PSE}

\begin{document}
\title{A Causal Bayesian Networks Viewpoint on Fairness}
%
%
\author{Silvia Chiappa\inst{}\and William S. Isaac\inst{}
\protect\footnotetext{The final publication is available at  \href{https://doi.org/10.1007/978-3-030-16744-8_1}{doi.org/10.1007/978-3-030-16744-8{\_}1} \cite{chiappa19causal}.}
}
\authorrunning{S. Chiappa and W. S. Isaac}
%
\institute{DeepMind, London, UK\\ 
\email{\{csilvia,williamis\}@google.com}}

\maketitle              
\begin{abstract}
We offer a graphical interpretation of unfairness in a dataset as the presence of an unfair causal path in the causal Bayesian network representing the data-generation mechanism. We use this viewpoint to revisit the recent debate surrounding the COMPAS pretrial risk assessment tool and, more generally, to point out that fairness evaluation on a model requires careful considerations on the patterns of unfairness underlying the training data. 
We show that causal Bayesian networks provide us with a powerful tool to measure unfairness in a dataset and to design fair models in complex unfairness scenarios.
\end{abstract}

\section{Introduction\label{sec:Intro}}
Machine learning is increasingly used in a wide range of decision-making scenarios that have serious implications for individuals and society, including financial lending \cite{byanjankar15predicting,malekipirbazari15risk}, hiring \cite{bogen18hiring,hoffman18discretion}, online advertising \cite{he2014practical,perlich14machine}, pretrial and immigration detention \cite{angwin16machine,rosenberg18immigration}, child maltreatment screening \cite{chouldechova18child,vaithianathan13child}, health care \cite{defauw18clinically,kourou15machine}, and social services \cite{ainowinstitute18litigating,eubanks18automating}. 
Whilst this has the potential to overcome undesirable aspects of human decision-making, there is concern that biases in the training data and model inaccuracies can lead to decisions that treat historically discriminated groups unfavourably. The research community has therefore started to investigate how to ensure that learned models do not take decisions that are \emph{unfair} with respect to \emph{sensitive attributes} (\eg~race or gender). 

This effort has led to the emergence of a rich set of fairness definitions \cite{chouldechova17fair,corbett-davies17algorithmic,dwork12fairness,feldman15certifying,mitchell18fair} providing researchers and practitioners with criteria to evaluate existing systems or to design new ones. Many such definitions have been found to be mathematically incompatible 
\cite{berk18fairness,chouldechova17fair,corbett-davies16computer,corbett-davies17algorithmic,kleinberg16inherent},
and this has been viewed as representing an
unavoidable trade-off establishing fundamental limits on fair machine learning, or as an indication that certain definitions do not map on to social or legal understandings of fairness \cite{corbett-davies18measure}. 

Most fairness definitions express properties of the model output with respect to sensitive information, without considering the relations among the relevant variables underlying the data-generation mechanism. As different relations would require a model to satisfy different properties in order to be fair, this could lead to erroneously classify as fair/unfair models exhibiting undesirable/legitimate biases.

In this manuscript, we use the causal Bayesian network framework to draw attention to this point, by visually describing unfairness in a dataset as the presence of an unfair \emph{causal path} in the data-generation mechanism. We then use this viewpoint to raise concern on the fairness debate surrounding the COMPAS pretrial risk assessment tool. Finally, we show that causal Bayesian networks offer a powerful tool for representing, reasoning about, and dealing with complex unfairness scenarios.

\section{A Graphical View of (Un)fairness\label{sec:SDF}}
Consider a dataset $\Delta =\{a^n,x^n,y^n\}_{n=1}^N$, corresponding to $N$ individuals, where $a^n$ indicates a sensitive attribute, and $x^n$ a set of observations that can be used (possibly together with $a^n$) to form a prediction $\hat y^n$ of outcome $y^n$. 
We assume a binary setting $a^n,y^n,\hat y^n\in\{0,1\}$ (unless otherwise specified), and indicate with $A,{\cal X}$, $Y$, and $\hat Y$ the (set of) random variables\footnote{Throughout the paper, we use capital and small letters for random variables and their values, and calligraphic capital letters for sets of variables.} corresponding to $a^n,x^n,y^n$, and $\hat y^n$ respectively.

In this section we show at a high-level that a correct use of fairness definitions concerned with statistical properties of $\hat Y$ with respect to $A$ requires an understanding of the patterns of unfairness underlying $\Delta$, and therefore of the relationships among $A$, ${\cal X}$ and $Y$. More specifically we show that:
\vskip-0.5cm
\begin{description}
\item[(i)] Using the framework of causal Bayesian networks (\CBNs), unfairness in $\Delta$ can be viewed as the presence of an unfair causal path from $A$ to ${\cal X}$ or $Y$.\\[-8pt] 
\item[(ii)] In order to determine which properties $\hat Y$ should possess to be fair, it is necessary to question and understand unfairness in $\Delta$.
\end{description}

\begin{wrapfigure}[5]{l}{0.2\textwidth}
\vskip-0.75cm
\scalebox{0.78}{
\begin{tikzpicture}[dgraph]
\node[ocont] (A) at (0,1.5) {$A$};
\node[ocont] (Q) at (2,1.5) {$Q$};
\node[ocont] (D) at (0,0) {$D$};
\node[ocont] (Y) at (2,0) {$Y$};
\draw[line width=1.15pt, postaction={draw,red,dash pattern= on 3pt off 6pt,dash phase=4pt}][line width=1.15pt, black,dash pattern= on 3pt off 6pt] (A)--node[sloped,above]{fair?}++(D);
\draw[line width=1.15pt,red](A)--node[sloped,above,black]{unfair}++(Y);
\draw[line width=1.15pt, postaction={draw,red,dash pattern= on 3pt off 6pt,dash phase=4pt}][line width=1.15pt, black,dash pattern= on 3pt off 6pt] (D)--node[sloped,above]{fair?}(Y);
\draw[line width=1.15pt](Q)--(Y);
\end{tikzpicture}}
\end{wrapfigure}
\noindent Assume a dataset $\Delta =\{a^n,x^n=\{q^n,d^n\},y^n\}_{n=1}^N$ corresponding to a college admission scenario in which applicants are admitted based on qualifications $Q$, choice of department $D$, and gender $A$; and in which female applicants apply more often to certain departments. This scenario can be represented by the \CBN~on the left (see \appref{sec:BN} for an overview of BNs, and \secref{sec:CBN} for a detailed treatment of \CBNs). The causal path $A\rightarrow Y$ represents direct influence of gender $A$ on admission $Y$, capturing the fact that two individuals with the same qualifications and applying to the same department can be treated differently depending on their gender. The indirect causal path $A\rightarrow D \rightarrow Y$ represents influence of $A$ on $Y$ through $D$, capturing the fact that female applicants more often apply to certain departments. Whilst the direct path $A\rightarrow Y$ is certainly an unfair one, the paths $A\rightarrow D$ and $D\rightarrow Y$, and therefore $A\rightarrow D \rightarrow Y$, could either be considered as fair or as unfair. For example, rejecting women more often due to department choice could be considered fair with respect to college responsibility. However, this could be considered unfair with respect to societal responsibility if the departmental differences were a result of systemic historical or cultural factors (\eg~if female applicants apply to specific departments at lower rates because of overt or covert societal discouragement). Finally, if the college were to lower the admission rates for departments chosen more often by women, then the path $D \rightarrow Y$ would be unfair.

Deciding whether a path is fair or unfair\footnote{A path could also be only partially fair --- we omit this case for simplicity.} requires careful ethical and sociological considerations and/or might not be possible from a dataset alone. Nevertheless, this example illustrates that we can view unfairness in a dataset as the presence of an unfair causal path from the sensitive attribute $A$ to ${\cal X}$ or $Y$. 

Different (un)fair path labeling requires $\hat Y$ to have different characteristics in order to be fair. 
In the case in which the causal paths from $A$ to $Y$ are all unfair (\eg~if $A\rightarrow D \rightarrow Y$ is considered unfair), a $\hat Y$ that is statistically independent of $A$ (denoted with $\hat Y \independent A $) would not contain any of the unfair influence of $A$ on $Y$. In such a case, $\hat Y$ is said to satisfy \emph{demographic parity}.\\

\noindent {\bf Demographic Parity (DP).} 
$\hat Y$ satisfies demographic parity if $\hat Y \independent A $, \ie~$p(\hat Y=1|A=0)=p(\hat Y=1|A=1)$, where \eg~$p(\hat Y=1|A=0)$ can be estimated as 
\begin{align*}
p(\hat Y=1|A=0)
\approx \frac{1}{N_0} \sum_{n=1}^{N} \mathbb{1}_{\hat y^n = 1, a^n=0}\,,
\end{align*}
with $\mathbb{1}_{\hat y^n = 1, a^n=0}=1$ if $\hat y^n = 1$ and $a^n=0$ (and zero otherwise), and where $N_0$ indicates the number of individuals with $a^n=0$. Notice that many classifiers, rather than a  binary prediction $\hat y^n\in\{0,1\}$, output a degree of belief that the individual belongs to class 1, $r^n$, also called \emph{score}. For example, in the case of logistic regression $r^n$ corresponds to the probability of class 1, \ie~$r^n=p(Y = 1 | a^n, x^{n})$. To obtain the prediction $\hat y^n\in\{0,1\}$ from $r^n$, it is common to use a threshold $\theta$, \ie~$\hat y^n=\mathbb{1}_{r^n>\theta}$. In this case, we can rewrite the estimate for $p(\hat Y=1|A=0)$ as 
\begin{align*}
p(\hat Y=1|A=0) \approx \frac{1}{N_0}\sum_{n=1}^{N}\mathbb{1}_{r^n>\theta,a^n=0}\,.
\end{align*}
Notice that $R \independent A$ implies $\hat Y \independent A$ for all values of $\theta$.\\

In the case in which the causal paths from $A$ to $Y$ are all fair (\eg~if $A\rightarrow Y$ is absent and $A\rightarrow D\rightarrow Y$ is considered fair), a $\hat Y$ such that $\hat Y \independent A | Y$ or $Y \independent A | \hat Y$ would be allowed to contain such a fair influence, but the (dis)agreement between $Y$ and $\hat Y$ would not be allowed to depend on $A$. In these cases, $\hat Y$ is said to satisfy \emph{equal false positive/false negative rates} and \emph{calibration} respectively.\\

\noindent {\bf Equal False Positive and Negative Rates (EFPRs/EFNRs).} $\hat Y$ satisfies EFPRs and EFNRs if $\hat Y \independent A | Y$, \ie~ (EFPRs) $p(\hat Y=1|Y=0,A=0)=p(\hat Y=1|Y=0,A=1)$ and (EFNRs) $p(\hat Y=0|Y=1,A=0)=p(\hat Y=0|Y=1,A=1)$.\\ 

\noindent {\bf Calibration.} 
$\hat Y$ satisfies calibration if $Y \independent A | \hat Y$. In the case of score output $R$, this condition is often instead called \emph{predictive parity} at threshold $\theta$, $p(Y=1|R>\theta,A=0)=p(Y=1|R>\theta,A=1)$, and calibration defined as requiring $Y \independent A |R$.\\

In the case in which at least one causal path from $A$ to $Y$ is unfair (\eg~if $A\rightarrow Y$ is present), 
EFPRs/EFNRs and calibration are inappropriate criteria, as they would not require the unfair influence of $A$ on $Y$ to be absent from $\hat Y$ (\eg~a perfect model ($\hat Y = Y$) would automatically satisfy EFPRs/EFNRs and calibration, but would contain the unfair influence). This observation is particularly relevant to the recent debate surrounding the \emph{correctional offender management profiling for alternative sanctions} (COMPAS) pretrial risk assessment tool. We revisit this debate in the next section.

\subsection{The COMPAS Debate \label{sec:COMPAS}}
Over the past few years, numerous state and local governments around the United States have sought to reform their pretrial court systems with the aim of reducing unprecedented levels of incarceration, and specifically the population of low-income defendants and racial minorities in America's prisons and jails \cite{alexander12new,flores16fpr,koepke17danger}. As part of this effort, quantitative tools for determining a person's likelihood for reoffending or failure to appear, called \emph{risk assessment instruments} (RAIs), were introduced to replace previous systems driven largely by opaque discretionary decisions and money bail \cite{arnold18racial,hls18bail}. However, the expansion of pretrial RAIs has unearthed new concerns of racial discrimination which would nullify the purported benefits of these systems and adversely impact defendants' civil liberties.

An intense ongoing debate, in which the research community has also been heavily involved, was triggered by an expos\'e from investigative journalists at ProPublica \cite{angwin16machine} on the COMPAS pretrial RAI developed by Equivant (formerly Northpointe) and deployed in Broward County in Florida. 
The COMPAS \emph{general recidivism risk scale} (GRRS) and \emph{violent recidivism risk scale} (VRRS), the focus of ProPublica’s investigation, sought to leverage machine learning techniques to improve the predictive accuracy of recidivism compared to older RAIs such as the level of service inventory-revised \cite{andrews00level} which were primarily based on theories and techniques from a sub-field of psychology known as the psychology of criminal conduct \cite{andrews06rai,brennan09evaluating}\footnote{While the exact methodology underlying GRRS and VRRS is proprietary, publicly available reports suggest that the process begins with a defendant being administered a 137 point assessment during intake. This is used to create a series of dynamic risk factor scales such as the \emph{criminal involvement scale} and \emph{history of violence scale}. In addition, COMPAS also includes static attributes such as the defendant’s age and prior police contact (number of prior arrests). The raw COMPAS scores are transformed into decile values by ranking and calibration with a normative group to ensure an equal proportion of scores within each scale value. Lastly, to aid practitioner interpretation, the scores are grouped into three risk categories. The scale values are displayed to court officials as either Low (1-4), Medium (5-7), and High (8-10) risk.}.
\begin{figure}[t]
\centering
\includegraphics[height=2.2in, width=4.9in]{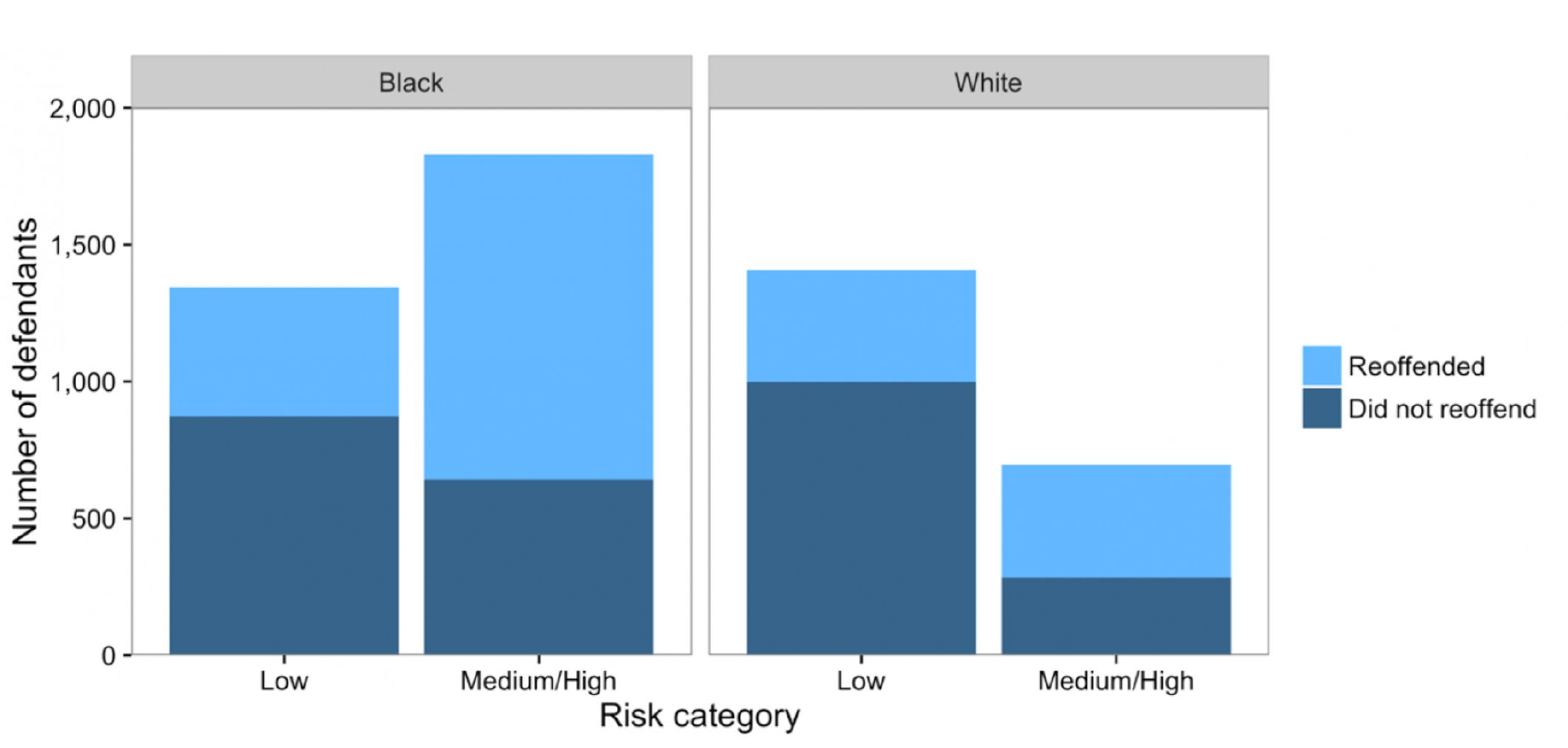}
\caption{Number of black and white defendants in each of two aggregate risk categories \cite{corbett-davies16computer}. The overall recidivism rate for black defendants is higher than for white defendants (52\% vs. 39\%), \ie~$Y\cancel{\independent} A$. Within each risk category, the proportion of defendants who reoffend is approximately the same regardless of race, \ie~$Y\independent A|\hat Y$. Black defendants are more likely to be classified as medium or high risk (58\% vs. 33\%) \ie~$\hat Y\cancel{\independent} A$. Among individuals who did not reoffend, black defendants are more likely to be classified as medium or high risk than white defendants (44.9\% to 23.5\%). Among individuals who did reoffend, white defendant are more likely to be classified as low risk than black defendants (47.7\% vs 28\%), \ie~$\hat Y\cancel{\independent} A| Y$.}
\label{fig:COMPAS}
\end{figure}

ProPublica’s criticism of COMPAS centered on two concerns. First, the authors argued that the distribution of the risk score $R\in\{1,\ldots,10\}$ exhibited discriminatory patterns, as black defendants displayed a fairly uniform distribution across each value, while white defendants exhibited a right skewed distribution, suggesting that the COMPAS recidivism risk scores disproportionately rated white defendants as lower risk than black defendants. Second, the authors claimed that the GRRS and VRRS did not satisfy EFPRs and EFNRs, as $\textrm{FPRs }=44.9\%$ and $\textrm{FNRs }=28.0\%$ for black defendants, whilst $\textrm{FPRs }=23.5\%$ and $\textrm{FNRs }=47.7\%$ for white defendants (see \figref{fig:COMPAS}). 
This evidence led ProPublica to conclude that COMPAS had a disparate impact on black defendants, leading to public outcry over potential biases in RAIs and machine learning writ large.

In response, Equivant published a technical report \cite{dieterich16compas} refuting the claims of bias made by ProPublica and concluded that COMPAS is sufficiently calibrated, in the sense that it satisfies  predictive parity at key thresholds. 
Subsequent analyses \cite{chouldechova17fair,corbett-davies17algorithmic,kleinberg16inherent} confirmed Equivant’s claims of calibration, but also demonstrated the incompatibility of EFPRs/EFNRs and calibration due to differences in base rates across groups ($Y\cancel{\independent} A$) (see \appref{sec:EFPRFNRC}). Moreover, the studies suggested that attempting to satisfy these competing forms of fairness force unavoidable trade-offs between criminal justice reformers’ purported goals of racial equity and public safety.

As explained in \secref{sec:SDF}, by requiring the rate of (dis)agreement between $Y$ and $\hat Y$ to be the same for black and white defendants (and therefore by not being concerned with dependence of $Y$ on $A$), EFPRs/EFNRs and calibration are inappropriate fairness criteria if dependence of $Y$ on $A$ includes influence of $A$ on $Y$ through an unfair causal path.  

\begin{wrapfigure}[9]{l}{0.26\textwidth}
\vskip-0.6cm
\centering
\scalebox{0.78}{
\begin{tikzpicture}[dgraph]
\node[ocont] (A) at (0,1.8) {$A$};
\node[ocont] (M) at (2,1.8) {${\mathcal M}$};
\node[ocont] (X) at (0,0) {${\mathcal F}$};
\node[ocont] (Y) at (2,0) {$Y$};
\draw[line width=1.15pt, postaction={draw,red,dash pattern= on 3pt off 6pt,dash phase=4pt}][line width=1.15pt, black,dash pattern= on 3pt off 6pt](A)--(X);\draw[line width=1.15pt,red](A)--(Y);\draw[line width=1.15pt](M)--(Y);
\draw[line width=1.15pt](X)--(Y);
\end{tikzpicture}}
\vspace{-0.3cm}
\caption{Possible \CBN\newline underlying the dataset used for COMPAS.}
\label{fig:CBNCOMPAS}
\end{wrapfigure}
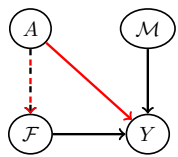
\noindent As previous research has shown \cite{isaac17hope,lum16predict,selbst17disparate}, modern policing tactics center around targeting a small number of neighborhoods --- often disproportionately populated by non-white and low income residents --- with recurring patrols and stops. This uneven distribution of police attention, as well as other factors such as funding for pretrial services \cite{koepke17danger,stevenson17assessing}, means that differences in base rates between racial groups are not reflective of ground truth rates. We can rephrase these findings as indicating the presence of a direct path $A\rightarrow Y$ (through unobserved neighborhood) in the \CBN~representing the data-generation mechanism (\figref{fig:CBNCOMPAS}). Such tactics also imply an influence of $A$ on $Y$ through the set of variables ${\mathcal F}$ containing number of prior arrests. In addition, the influence of $A$ on $Y$ through $A\rightarrow Y$ and $A \rightarrow {\mathcal F} \rightarrow Y$ could be more prominent or contain more unfairness due to racial discrimination. 

These observations indicate that EFPRs/EFNRs and calibration are inappropriate criteria for this case (and therefore that their incompatibility is irrelevant), and more generally that the current fairness debate surrounding COMPAS gives insufficient consideration to the patterns of unfairness underlying the training data. Our analysis formalizes the concerns raised by social scientists and legal scholars on mismeasurement and unrepresentative data in the US criminal justice system. Multiple studies \cite{eckhouse18layer,lum17nature,mayson19bias,stevenson17assessing} have argued that the core premise of RAIs, to assess the likelihood a defendant reoffends, is impossible to measure and that the empirical proxy used (\eg~arrest or conviction) introduces embedded biases and norms which render existing fairness tests unreliable.\\

This section used the \CBN~framework to describe at a high-level different patterns of unfairness that can underlie a dataset and to point out issues with current deployment of fairness definitions. In the remainder of the manuscript, we use this framework more extensively to further advance our analysis on fairness. Before doing that, we give some background on \CBNs~\cite{dawid07fundamentals,pearl00causality,pearl16causal,peters16elements,spirtes00causation}, assuming that all variables except $A$ are continuous. 

\section{Causal Bayesian Networks\label{sec:CBN}}
A \emph{Bayesian network} is a \emph{directed acyclic graph} where nodes and edges represent random variables and statistical dependencies. Each node $X_i$ in the graph is associated with the conditional distribution $p(X_i|\textrm{pa}(X_i))$, where $\textrm{pa}(X_i)$ is the set of \emph{parents} of $X_i$. The joint distribution of all nodes, $p(X_1, \ldots, X_I)$, is given by the product of all conditional distributions, \ie~$p(X_1,\ldots,X_I)=\prod_{i=1}^Ip(X_i|\textrm{pa}(X_i))$ (see \appref{sec:BN} for more details on Bayesian networks).

When equipped with causal semantic, namely when representing the data-generation mechanism, Bayesian networks can be used to visually express causal relationships. More specifically, \CBNs~enable us to give a graphical definition of causes and causal effects: if there exists a \emph{directed  path} from $A$ to $Y$, then $A$ is a \emph{potential cause} of $Y$. Directed paths are also called \emph{causal paths}. 

\begin{wrapfigure}[12]{l}{0.5\textwidth}
\vspace{-0.8cm}
\centering
\subfigure[]{
\scalebox{0.78}{
\begin{tikzpicture}[dgraph]
\node at (1,0.6) {${\cal G}$}; 
\node[ocont] (C) at (1,1.5) {$C$};
\node[ocont] (A) at (0,0) {$A$};
\node[ocont] (Y) at (2,0) {$Y$};
\node[] at (1,2.1) {$p(C)$};
\node[] at (0,-0.7) {$p(A|C)$};
\node[] at (2,-0.7) {$p(Y|C,A)$};
\draw[line width=1.15pt](C)--(A);\draw[line width=1.15pt](C)--(Y);\draw[line width=1.15pt](A)--(Y);
\end{tikzpicture}}}
\subfigure[]{
\scalebox{0.78}{
\begin{tikzpicture}[dgraph]
\node at (0.9,0.6) {${\cal G}_{\rightarrow A=a}$}; 
\node[ocont] (C) at (1,1.5) {$C$};
\node[ocont] (A) at (0,0) {$A$};
\node[ocont] (Y) at (2,0) {$Y$};
\node[] at (1,2.1) {$p(C)$};
\node[] at (0,-0.7) {$\delta_{A=a}$};
\node[] at (2,-0.7) {$p(Y|C,A)$};
\draw[line width=1.15pt](C)--(Y);\draw[line width=1.15pt](A)--(Y);
\end{tikzpicture}}}
\vspace{-0.35cm}
\caption{(a): \CBN~with a confounder $C$ for the effect of $A$ on $Y$. (b): Modified \CBN~resulting from intervening on $A$.}
\label{fig:CE}
\end{wrapfigure}
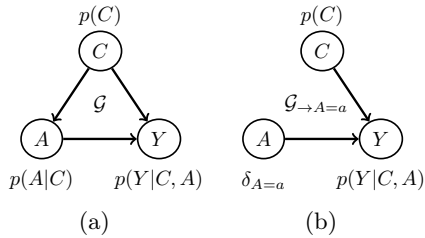
\noindent The causal effect of $A$ on $Y$ can be seen as the information traveling from $A$ to $Y$ through causal paths, or as the conditional distribution of $Y$ given $A$ restricted to causal paths. This implies that, to compute the causal effect, we need to disregard the information that travels along non-causal paths, which occurs if such paths are \emph{open}. Since paths with an arrow emerging from $A$ are either causal or closed (\emph{blocked}) by a \emph{collider}, the problematic paths are only those with an arrow pointing into $A$, called \emph{back-door paths}, which are open if they do not contain a collider.

An example of an open back-door path is given by $A\leftarrow C \rightarrow Y$ in the \CBN~${\cal G}$ of \figref{fig:CE}(a): 
the variable $C$ is said to be a \emph{confounder} for the effect of $A$ on $Y$, as it confounds the causal effect with non-causal information. 
To understand this, assume that $A$ represents hours of exercise in a week, $Y$ cardiac health, and $C$ age: observing cardiac health conditioning on exercise level from $p(Y|A)$ does not enable us to understand the effect of exercise on cardiac health, since $p(Y|A)$ includes the dependence between $A$ and $Y$ induced by age.

Each parent-child relationship in a \CBN~represents an autonomous mechanism, and therefore it is conceivable to change one such a relationship without changing the others. This enables us to express the causal effect of $A=a$ on $Y$
as the conditional distribution $p_{\rightarrow A=a}(Y|A=a)$ on the modified \CBN~${\cal G}_{\rightarrow A=a}$ of \figref{fig:CE}(b), resulting from replacing $p(A|C)$ with a Dirac delta distribution $\delta_{A=a}$ (thereby removing the link from $C$ to $A$) and leaving the remaining conditional distributions $p(Y|A,C)$ and $p(C)$ unaltered --- this process is called \emph{intervention} on $A$.
The distribution $p_{\rightarrow A=a}(Y|A=a)$ can be estimated as $p_{\rightarrow A=a}(Y|A=a) = \int_C p_{\rightarrow A=a}(Y|A=a,C)p_{\rightarrow A=a}(C|A=a) = \int_C p(Y|A=a,C)p(C)$. This is a special case of the following back-door adjustment formula.

{\bf Back-door Adjustment.} If a set of variables ${\cal C}$ satisfies the back-door criterion relative to $\{A, Y\}$, the causal effect of $A$ on $Y$ is given by $p_{\rightarrow A}(Y|A)=\int_{\cal C} p(Y|A,{\cal C})p({\cal C})$.
${\cal C}$ satisfies the back-door criterion if (a) no node in ${\cal C}$ is a \emph{descendant} of $A$
and (b) ${\cal C}$ blocks every back-door path from $A$ to $Y$.

The equality $p_{\rightarrow A=a}(Y|A=a,{\cal C}) = p(Y|A=a,{\cal C})$ follows from the fact that ${\cal G}_{A \rightarrow}$, obtained by removing from ${\cal G}$ all links emerging from $A$, retains all (and only) the back-door paths from $A$ to $Y$. As ${\cal C}$ blocks all such paths, $Y\independent A|{\cal C}$ in 
${\cal G}_{A \rightarrow}$. This means that there is no non-causal information traveling from $A$ to $Y$ when conditioning on ${\cal C}$ and therefore conditioning on $A$ coincides with intervening.

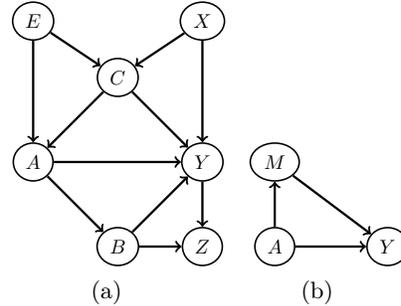
\begin{wrapfigure}[17]{r}{0.5\textwidth}
\vskip-0.5cm
\subfigure[]{
\scalebox{0.75}{
\begin{tikzpicture}[dgraph]
\node[ocont] (E) at (2,2.5) {$E$};
\node[ocont] (C) at (3.5,1.5) {$C$};
\node[ocont] (X) at (5,2.5) {$X$};
\node[ocont] (A) at (2,0) {$A$};
\node[ocont] (Y) at (5,0) {$Y$};
\node[ocont] (B) at (3.5,-1.5) {$B$};
\node[ocont] (Z) at (5,-1.5) {$Z$};
\draw[line width=1.15pt](E)--(A);\draw[line width=1.15pt](E)--(C);\draw[line width=1.15pt](X)--(C);\draw[line width=1.15pt](X)--(Y);
\draw[line width=1.15pt](C)--(A);\draw[line width=1.15pt](C)--(Y);\draw[line width=1.15pt](A)--(B);\draw[line width=1.15pt](B)--(Y);\draw[line width=1.15pt](A)--(Y);
\draw[line width=1.15pt](B)--(Z);\draw[line width=1.15pt](Y)--(Z);
\end{tikzpicture}}}
\subfigure[]{
\scalebox{0.75}{
\begin{tikzpicture}[dgraph]
\node[ocont] (M) at (0,1.5) {$M$};
\node[ocont] (X) at (0,0) {$A$};
\node[ocont] (Y) at (2,0) {$Y$};
\draw[line width=1.15pt](X)--(M);\draw[line width=1.15pt](X)--(Y);\draw[line width=1.15pt](M)--(Y);
\end{tikzpicture}}}
\vspace{-0.0cm}
\caption{(a): \CBN~in which conditioning on $C$ closes the paths $A\leftarrow C\leftarrow X \rightarrow Y$ and $A\leftarrow C\rightarrow Y$ but opens the path $A\leftarrow E\rightarrow C\leftarrow X \rightarrow Y$. (b): \CBN~with one direct and one indirect causal path from $A$ to $Y$.}
\label{fig:CBN}
\end{wrapfigure}
Conditioning on $C$ to block an open back-door path may open a closed path on which $C$ is a collider. For example, in the \CBN~of \figref{fig:CBN}(a), conditioning on $C$ closes the paths $A\leftarrow C\leftarrow X \rightarrow Y$ and $A\leftarrow C\rightarrow Y$,
but opens the path $A\leftarrow E\rightarrow C\leftarrow X \rightarrow Y$ (additional conditioning on $X$ would close $A\leftarrow E\rightarrow C\leftarrow X \rightarrow Y$).

The back-door criterion can also be derived from the rules of do-calculus \cite{pearl00causality,pearl16causal}, which indicate whether and how $p_{\rightarrow A}(Y|A)$ can be estimated using observations from ${\cal G}$: for many graph structures with unobserved confounders the only way to compute causal effects is by collecting observations directly from ${\cal G}_{\rightarrow A}$ --- in this case the effect is said to be \emph{non-identifiable}. 

\subsubsection*{Potential Outcome Viewpoint.}
Let $Y_{A=a}$ be the random variable with distribution $p(Y_{A=a}) = p_{\rightarrow A=a}(Y|A=a)$. $Y_{A=a}$ is called \emph{potential outcome} and, when not ambiguous, we will refer to it with the shorthand $Y_a$. The relation between $Y_{a}$ and all the variables in ${\cal G}$ other than $Y$ can be expressed by the graph  
obtained by removing from ${\cal G}$ all the links emerging from $A$, and by replacing $Y$ with $Y_{a}$. If $Y_{a}$ is independent on $A$ in this graph, then\footnote{The equality $p(Y_a|A=a)=p(Y|A=a)$ is called \emph{consistency}.} $p(Y_a)=p(Y_a|A=a)=p(Y|A=a)$. If $Y_{a}$ is independent of $A$ in this graph when conditioning on ${\cal C}$, then 
\begin{align*}
p(Y_{a}) = \int_{{\cal C}} p(Y_{a}|{\cal C}) p({\cal C}) = \int_{{\cal C}} p(Y_{a}|A=a,{\cal C}) p({\cal C}) = \int_{{\cal C}} p(Y|A=a,{\cal C}) p({\cal C}) \,,
\end{align*}
\ie~we retrieve the back-door adjustment formula.\\

In the remainder of the section we show that, by performing different interventions on $A$ along different causal paths, it is possible to isolate the contribution of the causal effect of $A$ on $Y$ along a group of paths. 

\subsection*{Direct and Indirect Effect}
Consider the \CBN~of \figref{fig:CBN}(b), containing the direct path $A\rightarrow Y$ and one indirect causal path through the variable $M$. Let $Y_{a}(M_{\bar a})$ be the 
random variable with distribution equal to the conditional distribution of $Y$ given $A$ restricted to causal paths, with $A=a$ along $A\rightarrow Y$ and $A=\bar a$ along $A\rightarrow M\rightarrow Y$. 
The \emph{average direct effect} (ADE) of $A=a$ with respect to $A=\bar a$, defined as
\begin{align*}
&\textrm{ADE}_{\bar a a } =\av{Y_{a}(M_{\bar a})}_{p(Y_{a}(M_{\bar a}))} - \av{Y_{\bar a}}_{p(Y_{\bar a})}\,,
\end{align*}
where \eg~$\av{Y_{\bar a}}_{p(Y_{\bar a})}=\int_{Y_{\bar a}} Y_{\bar a}p(Y_{\bar a})$, measures the difference in flow of causal information from $A$ to $Y$ between the case in which $A=a$ along $A\rightarrow Y$ and $A=\bar a$ along $A\rightarrow M\rightarrow Y$ and the case in which $A=\bar a$ along both paths. 

Analogously, the \emph{average indirect effect} (AIE) of $A=a$ with respect to $A=\bar a$, 
is defined as $\textrm{AIE}_{\bar a a } =\av{Y_{\bar a}(M_a)}_{p(Y_{\bar a}(M_a))} - \av{Y_{\bar a}}_{p(Y_{\bar a})}$.

The difference $\textrm{ADE}_{\bar a a } - \textrm{AIE}_{a \bar a }$ gives the \emph{average total effect} (ATE) $\textrm{ATE}_{\bar a a} = \av{Y_{a}}_{p(Y_{a})} - \av{Y_{\bar a}}_{p(Y_{\bar a})}$\footnote{Often the AIE of $A=a$ with respect to $A=\bar a$ is defined as $\textrm{AIE}^a_{\bar a a} = \av{Y_{a}}_{p(Y_{a})} - \av{Y_{a}(M_{\bar a})}_{p(Y_{a}(M_{\bar a}))}= -\textrm{AIE}_{a \bar a }$, which differs in setting $A$ to $a$ rather than to $\bar a$ along $A\rightarrow Y$. In the linear case, the two 
definitions coincide (see Eqs. (\ref{eq:AIE1}) and (\ref{eq:AIE2})).
Similarly the ADE can be defined as $\textrm{ADE}^a_{\bar a a} = \av{Y_{a}}_{p(Y_{a})} - \av{Y_{\bar a}(M_a)}_{p(Y_{\bar a}(M_a))}= -\textrm{ADE}_{a \bar a }$.}. 

\subsection*{Path-Specific Effect}
\begin{wrapfigure}[20]{r}{0.39\textwidth}
\vspace{-1.3cm}
\centering
\scalebox{0.78}{
\begin{tikzpicture}[dgraph]
\node[ocont] (C) at (3,2.5) {$C$};
\node[ocont] (A) at (0,1) {$A$};
\node[ocont] (M) at (1.5,1) {$M$};
\node[ocont] (L) at (3,1) {$L$};
\node[ocont] (Y) at (4.5,1) {$Y$};
\draw[line width=1.15pt,red](A)--(M);
\draw[line width=1.15pt,red] (M)--(L);
\draw[line width=1.15pt,red] (L)--(Y);
\draw[line width=1.15pt,red] (M)to [bend left=-30](Y);
\draw[line width=1.15pt](A)to [bend left=-30](L);
\draw[line width=1.15pt,red](A)to [bend right=+35](Y);
\draw[line width=1.15pt](C)--(M);
\draw[line width=1.15pt](C)--(L);
\draw[line width=1.15pt](C)--(Y);
\end{tikzpicture}}
\\
\scalebox{0.78}{
\begin{tikzpicture}[dgraph]
\node[ocont] (C) at (3.6,2.8) {$C$};
\node[ocont] (A) at (0,1) {$A$};
\node[ocont] (M) at (1.8,1) {$M$};
\node[ocont] (L) at (3.6,1) {$L$};
\node[ocont] (Y) at (5.4,1) {$Y$};
\draw[line width=1.15pt](A)--node[sloped,above]{$\theta^m_a$}++(M);
\draw[line width=1.15pt](M)--node[sloped,above]{$\theta^l_m$}++(L);
\draw[line width=1.15pt](L)--node[sloped,above]{$\theta^y_l$}++(Y);
\draw[line width=1.15pt](A)to [bend left=-48]node [above, sloped] (TextNode1) {$\theta^l_a$} (L);
\draw[line width=1.15pt](M)to [bend left=-48]node [above, sloped] (TextNode2) {$\theta^y_m$} (Y);
\draw [line width=1.15pt](A) to [bend right=48] node [above, sloped] (TextNode3) {$\theta^y_a$} (Y);
\draw[line width=1.15pt](C)--node[sloped,above]{$\theta^m_c$}++(M);
\draw[line width=1.15pt](C)--node[sloped,above,rotate=180]{$\theta^l_c$}++(L);
\draw[line width=1.15pt](C)--node[sloped,above]{$\theta^y_c$}++(Y);
\end{tikzpicture}}
\vspace{-0.4cm}
\caption{Top: \CBN~with the direct path from $A$ to $Y$ and the indirect paths passing through $M$ highlighted in red. Bottom: \CBN~corresponding to \eqref{eq:lm}.}
\label{fig:GCMPSE}
\end{wrapfigure}
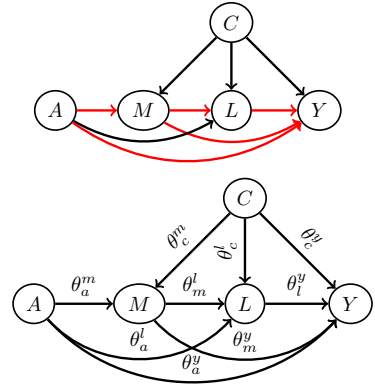
To estimate the effect along a specific group of causal paths, 
we can generalize the formulas for the ADE and AIE by replacing the variable in the first term with the one resulting from performing the intervention $A=a$ along the group of interest and $A=\bar a$ along the remaining causal paths. 
For example, consider the \CBN~of \figref{fig:GCMPSE} (top) and assume that we are interested in isolating the effect of $A$ on $Y$ along the direct path $A\rightarrow Y$ and the paths passing through $M$, $A\rightarrow M \rightarrow,\ldots,\rightarrow Y$, namely along the red links. The \emph{path-specific effect} (\PSE) of $A=a$ with respect to $A=\bar a$ for this group of paths is defined as
\begin{align*}
\textrm{PSE}_{\bar a a} = \av{Y_a(M_a, L_{\bar a}(M_a))}-\av{Y_{\bar a}}\,,
\end{align*}
where $p(Y_a(M_a, L_{\bar a} (M_a)))$ is given by 
\begin{align*}
\int_{C,M,L} p(Y|A=a,C,M,L)p(L|A=\bar a,C,M)p(M|A=a,C)p(C)\,.
\end{align*}
In the simple case in which the \CBN~corresponds to a linear model, \eg
\begin{align}
&A\sim\textrm{Bern}(\pi), \hskip0.1cm C = \epsilon_c\,,\nonumber\\
&M=\theta^m+\theta^m_{a}A+\theta^m_{c}C+\epsilon_m,\hskip0.1cm L=\theta^l+\theta^l_{a}A+\theta^l_{c}C+\theta^l_{m}M+\epsilon_l\,,\nonumber\\
&Y=\theta^y+\theta^y_{a}A+\theta^y_{c}C+\theta^y_{m}M+\theta^y_{l}L+\epsilon_y\,,\label{eq:lm}
\end{align}
where $\epsilon_c$, $\epsilon_m$, $\epsilon_l$ and $\epsilon_y$ are unobserved independent zero-mean Gaussian variables, 
we can compute $\av{Y_{\bar a}}$ by expressing $Y$ as a function of $A=\bar a$ and the Gaussian variables, by recursive substitutions in $C, M$ and $L$, \ie~
\begin{align*}
Y_{\bar a}&=\theta^y+\theta^y_{a}\bar a+\theta^y_{c}\epsilon_c+\theta^y_{m}(\theta^m+\theta^m_{a}\bar a+\theta^m_{c}\epsilon_c+\epsilon_m)\\
&\hskip0.32cm+\theta^y_{l}(\theta^l+\theta^l_{a}\bar a+\theta^l_{c}\epsilon_c+\theta^l_{m}(\theta^m+\theta^m_{a}\bar a+\theta^m_{c}\epsilon_c+\epsilon_m)+\epsilon_l)+\epsilon_y\,,
\end{align*}
and then take the mean, obtaining $\av{Y_{\bar a}}=\theta^y+\theta^y_{a}\bar a+\theta^y_{m}(\theta^m+\theta^m_{a}\bar a)+\theta^y_{l}(\theta^l+\theta^l_{a}\bar a+\theta^l_{m}(\theta^m+\theta^m_{a}\bar a))$. Analogously
\begin{align*}
\av{Y_a(M_a, L_{\bar a}(M_a))}&=\theta^y+\theta^y_{a}a+\theta^y_m(\theta^m+\theta^m_{a}a)+\theta^y_l(\theta^l+\theta^l_a\bar a+ \theta^l_m(\theta^m+\theta^m_{a}a))\,.
\end{align*}
For $a=1$ and $\bar a = 0$, this gives
\begin{align*}
\textrm{PSE}_{\bar a a}=\theta^y_{a}(a-\bar a)+\theta^y_{m}\theta^m_{a}(a-\bar a)+\theta^y_{l}\theta^l_{m}\theta^m_{a}(a-\bar a)=\theta^y_{a}+\theta^y_{m}\theta^m_{a}+\theta^y_{l}\theta^l_{m}\theta^m_{a}\,.
\end{align*}
The same conclusion could have been obtained by looking at the graph annotated with path coefficients (\figref{fig:GCMPSE} (bottom)). The PSE is obtained by summing over the three causal paths of interest ($A\rightarrow Y$, $A\rightarrow M \rightarrow Y$, and $A\rightarrow M \rightarrow L \rightarrow Y$) the product of all coefficients in each path.

Notice that $\textrm{AIE}_{\bar a a}$, given by
\begin{align}
\textrm{AIE}_{\bar a a} &= \av{Y_{\bar a}(M_a, L_a(M_a))}-\av{Y_{\bar a}}\nonumber\\
& = \theta^y+{\myblue \theta^y_{a}\bar a}+\theta^y_m(\theta^m+\theta^m_{a} a)+\theta^y_l(\theta^l+\theta^l_a a+ \theta^l_m(\theta^m+\theta^m_{a}a))\nonumber\\
& \hskip0.35cm - \theta^y+{\myblue \theta^y_{a}\bar a}+\theta^y_m(\theta^m+\theta^m_{a} \bar a)+\theta^y_l(\theta^l+\theta^l_a \bar a+ \theta^l_m(\theta^m+\theta^m_{a}\bar a))\nonumber\\
& = \theta^y_m\theta^m_{a} (a-\bar a) + \theta^y_l(\theta^l_a(a-\bar a)+ \theta^l_m\theta^m_{a}(a-\bar a))\,,
\label{eq:AIE1}
\end{align}
coincides with $\textrm{AIE}^a_{\bar a a}$, given by
\begin{align}
\textrm{AIE}^a_{\bar a a}& = \av{Y_{a}} - \av{Y_a(M_{\bar a}, L_{\bar a}(M_{\bar a}))}\nonumber\\
& = \theta^y+{\myblue \theta^y_{a}a}+\theta^y_m(\theta^m+\theta^m_{a} a)+\theta^y_l(\theta^l+\theta^l_a a+ \theta^l_m(\theta^m+\theta^m_{a}a))\nonumber\\
& \hskip0.35cm -\theta^y+{\myblue \theta^y_{a}a}+\theta^y_m(\theta^m+\theta^m_{a} \bar a)+\theta^y_l(\theta^l+\theta^l_a \bar a+ \theta^l_m(\theta^m+\theta^m_{a}\bar a))\,.
\label{eq:AIE2}
\end{align}

\subsubsection*{Effect of Treatment on Treated.}
Consider the conditional distribution $p(Y_{a}|A=\bar a)$. This distribution measures the information travelling from $A$ to $Y$ along all open paths, when $A$ is set to $a$ along causal paths and to $\bar a$ along non-causal paths. The \emph{effect of treatment on treated} (ETT) of $A=a$ with respect to $A=\bar a$ is defined as $\textrm{ETT}_{\bar a a} = \av{Y_{a}}_{p(Y_{a}|A=\bar a)}- \av{Y_{\bar a}}_{p(Y_{\bar a}|A=\bar a)} = \av{Y_{a}}_{p(Y_{a}|A=\bar a)} - \av{Y}_{p(Y|A=\bar a)}$. 
As the PSE, the ETT measures difference in flow of information from $A$ to $Y$ when $A$ takes different values along different paths. However, the PSE considers only causal paths and different values for $A$ along different causal paths, whilst the ETT considers all open paths and different values for $A$ along causal and non-causal paths respectively. 
Similarly to $\textrm{ATE}_{\bar a a}$, $\textrm{ETT}_{\bar a a}$ for the \CBN~of \figref{fig:CBN}(b) can be expressed as
\begin{align*}
\textrm{ETT}_{\bar a a} 
&=\underbrace{\av{Y_{a}(M_{\bar a})} - \av{Y_{\bar a}}}_{\textrm{ADE}_{\bar a a|\bar a}}
-(\underbrace{\av{Y_{a}(M_{\bar a})} - \av{Y_{a}}}_{\textrm{AIE}_{a \bar a|\bar a}})\,.
\end{align*}
Notice that, if we define difference in flow of non-causal (along the open back-door paths) information from $A$ to $Y$ when $A=a$ with respect to when $A=\bar a$ as $\textrm{NCI}_{\bar a a} = \av{Y_{\bar a}}_{p(Y_{\bar a}|A=a)} - \av{Y}_{p(Y|A=\bar a)}$, we obtain 
\begin{align*}
\av{Y}_{p(Y|A=a)} - \av{Y}_{p(Y|A=\bar a)} &= \av{Y_{\bar a}}_{p(Y_{\bar a}|A=a)} - \av{Y}_{p(Y|A=\bar a)}\\
& - (\av{Y_{\bar a}}_{p(Y_{\bar a}|A=a)} - \av{Y}_{p(Y|A=a)})\\
&= \textrm{NCI}_{\bar a a} - \textrm{ETT}_{a \bar a} = \textrm{NCI}_{\bar a a} - \textrm{ADE}_{a \bar a| a} + \textrm{AIE}_{\bar a a|a}\,.
\end{align*}

\section{Fairness Considerations using \CBNs}
Equipped with the background on \CBNs~from \secref{sec:CBN}, in this section we further investigate unfairness in a dataset $\Delta=\{a^n, x^n, y^n\}_{n=1}^N$, discuss issues that might arise when building a decision system from it, and show how to measure and deal with unfairness in complex scenarios, revisiting and extending material from \cite{chiappa19path,kusner17counterfactual,zhang18fairness}.

\subsection{Back-door Paths from $A$ to $Y$}
In \secref{sec:SDF} we have introduced a graphical interpretation of unfairness in a dataset $\Delta$ as the presence of an unfair causal path from $A$ to ${\cal X}$ or $Y$. More specifically, we have shown through a college admission example that unfairness can be due to an unfair link emerging (a) from $A$ or (b) from a subsequent variable in a causal path from $A$ to $Y$ (\eg~ $D\rightarrow Y$ in the example). Our discussion did not mention paths from $A$ to $Y$ with an arrow pointing into $A$, namely back-door paths. This is because such paths are not problematic.

\begin{wrapfigure}[5]{l}{0.22\textwidth}
\vskip-0.7cm
\centering
\scalebox{0.78}{
\begin{tikzpicture}[dgraph]
\node[ocont] (A) at (0,1.5) {$A$};
\node[ocont] (E) at (2,1.5) {$E$};
\node[ocont] (Y) at (2,0) {$Y$};
\draw[line width=1.15pt](E)--(A);
\draw[line width=1.15pt](E)--(Y);
\end{tikzpicture}}
\end{wrapfigure}
\noindent To understand this, consider the hiring scenario described by the \CBN~on the left, where $A$ represents religious belief and $E$ educational background of the applicant, which influences religious participation ($E\rightarrow A$). 
Whilst $Y \cancel{\independent} A$ due to the open back-door path from $A$ to $Y$, the hiring decision $Y$ is only based on $E$.

\subsection{Opening Closed Unfair Paths from $A$ to $Y$}
In \secref{sec:SDF}, we have seen that, in order to reason about fairness of $\hat Y$, it is necessary to question and understand unfairness in $\Delta$. In this section, we warn that another crucial element needs to be considered in the fairness discussion around $\hat Y$, namely  
\begin{description}
\item[(i)] The variables used to form $\hat Y$ could project into $\hat Y$ unfair patterns in ${\cal X}$ that do not concern $Y$. 
\end{description}
This could happen, for example, if a closed unfair path from $A$ to $Y$ is opened when conditioning on the variables used to form $\hat Y$. 

\begin{wrapfigure}[9]{l}{0.22\textwidth}
\vskip0.1cm
\centering
\scalebox{0.8}{
\begin{tikzpicture}[dgraph]
\node[ocont] (A) at (-0.9,1.5) {$A$};
\node[ocont] (M) at (0.9,1.5) {$M$};
\node[ocont] (X) at (0,0) {$X$};
\node[ocont] (Y) at (1.6,0) {$Y$};
\draw[line width=1.15pt,red](A)--node[sloped,above,black]{$\alpha$}++(X);
\draw[line width=1.15pt](M)--node[sloped,above]{$\beta$}++(X);
\draw[line width=1.15pt](M)--node[sloped,above]{$\gamma$}++(Y);
\end{tikzpicture}}
\vspace{-0.4cm}
\caption{\CBN~underlying a music degree scenario.}
\label{fig:TS}
\end{wrapfigure}
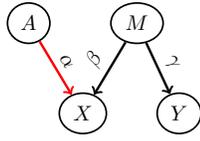
\noindent As an example, assume the \CBN~in \figref{fig:TS} representing the data-generation mechanism underlying a music degree scenario, where $A$ corresponds to gender, $M$ to music aptitude (unobserved, \ie~$M\notin \Delta$), $X$ to the score obtained from an ability test taken at the beginning of the degree, and $Y$ to the score obtained from an ability test taken at the end of the degree. Individuals with higher music aptitude $M$ are more likely to obtain higher initial and final scores ($M\rightarrow X$, $M\rightarrow Y$). Due to discrimination occurring at the initial testing, women are assigned a lower initial score than men for the same aptitude level ($A \rightarrow X$). The only path from $A$ to $Y$, $A\rightarrow X \leftarrow M \rightarrow Y$, is closed as $X$ is a collider on this path. Therefore the unfair influence of $A$ on $X$ does not reach $Y$ ($Y\independent A$). Nevertheless, as $Y\cancel{\independent} A |X$, a prediction $\hat Y$ based on the initial score $X$ only would contain the unfair influence of $A$ on $X$. For example, assume the following linear model: $Y=\gamma M, \hskip0.3cm X =\alpha A + \beta M$, with $\av{A^2}_{p(A)}=1$ and $\av{M^2}_{p(M)}=1$. A linear predictor of the form $\hat Y = \theta_X X$ minimizing $\av{(Y-\hat Y)^2}_{p(A)p(M)}$ would have parameters $\theta_X=\gamma\beta/(\alpha^2+\beta^2)$, 
giving $\hat Y = \gamma\beta(\alpha A + \beta M)/(\alpha^2+\beta^2)$, \ie~$\hat Y\cancel{\independent} A$.
Therefore, this predictor would be using the sensitive attribute to form a decision, although implicitly rather than explicitly. Instead, a predictor explicitly using the sensitive attribute, $\hat Y = \theta_X X + \theta_A A$, would have parameters
\begin{align*}
\left(\begin{array}{c}
\theta_X \\
\theta_A \\
\end{array} \right)
&=\left(
\begin{array}{cc}
\alpha^2+\beta^2 & \alpha \\
\alpha  & 1 \\
\end{array} \right)^{-1}
\left(\begin{array}{c}
\gamma\beta  \\
0\\
\end{array} \right)
=\left(\begin{array}{c}
\gamma/\beta \\
-\alpha\gamma/\beta \\
\end{array} \right),
\end{align*}
\ie~$\hat Y = \gamma M$.
Therefore, this predictor would be fair. From the \CBN~we can see that the explicit use of $A$ can be of help in retrieving $M$. Indeed, since $M\cancel{\independent} A |X$, using $A$ in addition to $X$ can give information about $M$. In general (\eg~in a non-linear setting) it is not guaranteed that using $A$ would ensure $\hat Y \independent A$. Nevertheless, this example shows how explicit use of the sensitive attribute in a model can ensure fairness rather than leading to unfairness. 

This observation is relevant to one of the simplest fairness definitions, motivated by legal requirements, called \emph{fairness through unawareness}, which states that $\hat Y$ is fair as long as it does not make explicit use of the sensitive attribute $A$. Whilst this fairness criterion is often indicated as problematic because some of the variables used to form $\hat Y$ could be a proxy for $A$ (such as neighborhood for race), the example above shows a more subtle issue with it.

\subsection{Path-Specific Population-level Unfairness}
In this section, we show that the path-specific effect introduced in \secref{sec:CBN} can be used to quantify unfairness in $\Delta$ in complex scenarios. 

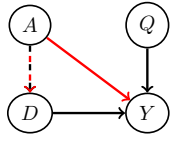
\begin{wrapfigure}[9]{l}{0.3\textwidth}
\vskip-0.4cm
\centering
\scalebox{0.78}{
\begin{tikzpicture}[dgraph]
\node[ocont] (A) at (0,1.5) {$A$};
\node[ocont] (C) at (2,1.5) {$Q$};
\node[ocont] (X) at (0,0) {$D$};
\node[ocont] (Y) at (2,0) {$Y$};
\draw[line width=1.15pt, postaction={draw,red,dash pattern= on 3pt off 6pt,dash phase=4pt}][line width=1.15pt, black,dash pattern= on 3pt off 6pt] (A)--(X);\draw[line width=1.15pt,red](A)--(Y);\draw[line width=1.15pt](X)--(Y);
\draw[line width=1.15pt](C)--(Y);
\end{tikzpicture}}
\vspace{-0.2cm}
\caption{\CBN~underlying a college admission scenario.}
\label{fig:CBNCA}
\end{wrapfigure}

Consider the college admission example discussed in \secref{sec:SDF} (\figref{fig:CBNCA}). In the case in which the path $A\rightarrow D$, and therefore $A\rightarrow D\rightarrow Y$, is considered unfair, unfairness overall population can be quantified with $\av{Y}_{p(Y|a)}-\av{Y}_{p(Y|{\bar a})}$ (coinciding with $\textrm{ATE}_{\bar a a} = \av{Y_{a}}_{p(Y_{a})}-\av{Y_{\bar a}}_{p(Y_{\bar a})}$) where, for example, $A=a$ and $A=\bar a$ indicate female and male applicants respectively. 

\noindent In the more complex case in which the path $A \rightarrow D\rightarrow Y$ is considered fair, unfairness can instead be quantified with the path-specific effect along the direct path $A\rightarrow Y$, $\textrm{PSE}_{\bar a a}$, given by 
\begin{align*}
\av{Y_{a}(D_{\bar a})}_{p(Y_{a}(D_{\bar a}))}-\av{Y_{\bar a}}_{p(Y_{\bar a})}\,.
\end{align*}
Notice that computing $p(Y_{a}(D_{\bar a}))$ requires knowledge of the \CBN. 
If the \CBN~structure is not known or estimating its conditional distributions is challenging, the resulting estimate could be imprecise.

\subsection{Path-Specific Individual-level Unfairness}
In the college admission example of \figref{fig:CBNCA} in which the path $A \rightarrow D\rightarrow Y$ is considered fair, rather than measuring unfairness overall population, we might want to know \eg~whether a rejected female applicant $\{a^n=a, q^n, d^n, y^n=0\}$ was treated unfairly. We can answer this question by estimating whether the applicant would have been admitted had she been male ($A=\bar a$) along the direct path $A\rightarrow Y$ from $p(Y_{\bar a}(D_a)|A=a, Q=q^n, D=d^n, Y=y^n)$ (notice that the outcome in the actual world, $y^n$, corresponds to $p(Y_{a}(D_a)|A=a, Q=q^n, D=d^n)=\mathbb{1}_{Y_{a}(D_a)=y^n}$). 

To understand how this can be achieved, consider the following linear model associated to a \CBN~with the same structure as the one in \figref{fig:CBNCA}
\begin{align}
&A\sim\textrm{Bern}(\pi), \hskip0.3cm Q=\theta^q+\epsilon_q, \hskip0.3cm D=\theta^d+\theta^d_{a}A+\epsilon_d\,,\nonumber\\ 
&\hskip0.0cm Y=\theta^y+\theta^y_{a}A+\theta^y_{q}Q+\theta^y_{d}D+\epsilon_y\,,
\label{eq:model}
\end{align}
where $\epsilon_q,\epsilon_d$ and $\epsilon_y$ are unobserved independent zero-mean Gaussian variables.

\begin{wrapfigure}[11]{l}{0.3\textwidth}
\vskip-0.8cm
\centering
\scalebox{0.85}{
\begin{tikzpicture}[dgraph]
\node[] (eD) at (1,2) {$\epsilon_d$};
\node[] (eY) at (1,0.5) {$\epsilon_y$};
\node[] (eQ) at (1,-1) {$\epsilon_q$};
\node[] (A) at (0,3) {$A$};
\node[] (D) at (0,1.5) {$D$};
\node[] (Y) at (0,0) {$Y$};
\node[] (Q) at (0,-1.5) {$Q$};
\node[] (D*) at (2.5,1.5) {$D_a$};
\node[] (Y*) at (2.5,0) {$Y_{\bar a}(D_a)$};
\node[] (Q*) at (2.5,-1.5) {$Q^*$};
\draw[line width=1.15pt](A)--(D);\draw[line width=1.15pt](A)to [bend left=-30](Y);\draw[line width=1.15pt](D)--(Y);
\draw[line width=1.15pt](Q)--(Y);
\draw[line width=1.15pt](Q*)--(Y*);\draw[line width=1.15pt](D*)--(Y*);
\draw[line width=1.15pt](eD)--(D);\draw[line width=1.15pt](eD)--(D*);
\draw[line width=1.15pt](eQ)--(Q);\draw[line width=1.15pt](eQ)--(Q*);
\draw[line width=1.15pt](eY)--(Y);\draw[line width=1.15pt](eY)--(Y*);
\end{tikzpicture}}
\end{wrapfigure}

\noindent The relationships between $A,Q,D,Y$ and $Y_{\bar a}(D_a)$ in this model can be inferred from the \emph{twin Bayesian network} \cite{pearl00causality} on the left resulting from the intervention $A=a$ along $A\rightarrow D$ and $A=\bar a$ along $A\rightarrow Y$: in addition to $A,Q,D$ and $Y$, the network contains the variables $Q^*$, $D_a$ and $Y_{\bar a}(D_a)$ corresponding to the counterfactual world in which $A=\bar a$ along $A\rightarrow Y$, with  $Q^*=\theta^q+\epsilon_q, \hskip0.0cm D_a=\theta^d+\theta^d_{a}a+\epsilon_d$, and $Y_{\bar a}(D_a)=\theta^y+\theta^y_a\bar a+\theta^y_{q}Q^*+\theta^y_{d}D_a +\epsilon_y$. The two groups of variables are connected through $\epsilon_d, \epsilon_q, \epsilon_y$, indicating that the factual and counterfactual worlds share the same unobserved randomness. From this network, we can deduce that $Y_{\bar a}(D_a)\independent \{A,Q,D,Y\}|\epsilon=\{\epsilon_q,\epsilon_d, \epsilon_y\}$\footnote{Notice that $Y_{\bar a}(D_a)\independent A$, but $Y_{\bar a}(D_a)\cancel{\independent} A|D$.}, and therefore that we can express $p(Y_{\bar a}(D_a)|A=a, Q=q^n, D=d^n, Y=y^n)$ as
\begin{align}
p(Y_{\bar a}(D_a)|a,q^n, d^n,y^n) 
= \int_{\epsilon} p(Y_{\bar a}(D_a)| \epsilon, \cancel{a}, \cancel{q^n}, \cancel{d^n}, \cancel{y^n})p(\epsilon|a, q^n, d^n, y^n)\,.
\label{eq:counterf}
\end{align}
As $p(\epsilon^n_q|a, q^n, d^n, y^n)=\delta_{\epsilon^n_q=q^n-\theta^q}$, $p(\epsilon^n_d|a, q^n, d^n, y^n)=\delta_{\epsilon^n_d=d^n-\theta^d-\theta^d_aa}$, and $p(\epsilon^n_y|a, q^n, d^n, y^n) = \delta_{\epsilon^n_y=y^n-\theta^y-\theta^y_aa-\theta^y_qq^n-\theta^y_dd^n}$, we obtain $$p(Y_{\bar a}(D_a)|A=a, Q=q^n, D=d^n, Y=y^n)=\mathbb{1}_{Y_{\bar a}(D_a)=y^n-\theta^y_aa+\theta^y_a\bar a}.$$
Indeed, by expressing $Y_{\bar a}(D_a)$ as a function of $\epsilon^n_q, \epsilon^n_d$ and $\epsilon^n_y$, we obtain
\begin{align*}
Y_{\bar a}(D_a)&=\theta^y+\theta^y_a\bar a+\theta^y_{q}Q^*+\theta^y_{d}D_a+\epsilon^n_y\\
& =\theta^y+\theta^y_a\bar a+\theta^y_{q}(\theta^q+\epsilon^n_q)+\theta^y_{d}(\theta^d+\theta^d_{a}a+\epsilon^n_d)+\epsilon^n_y=y^n-\theta^y_aa+\theta^y_a\bar a\,.
\end{align*}
Therefore, as $Q$ is not a descendant of $A$ and $D$ is a descendant of $A$ along a fair path, the outcome in the counterfactual world is obtained by correcting the outcome in the factual world through replacing $\theta^y_aa$ with $\theta^y_a\bar a$. 

Suppose that we want to post-process a learned model (\ref{eq:model}) to give a prediction $\hat y^n$ for  individual $\{a^n=a, q^n, d^n\}$ based on the counterfactual distribution $p(Y_{\bar a}(D_a)|A=a, Q=q^n, D=d^n)$, \eg~$\hat y^n=\av{Y_{\bar a}(D_a)}_{p(Y_{\bar a}(D_a)|A=a, Q=q^n, D=d^n)}$ --- the resulting model is said to satisfy \emph{path-specific counterfactual fairness} \cite{chiappa19path}. 
By performing a similar reasoning to the one above for $p(Y_{\bar a}(D_a)|A=a, Q=q^n, D=d^n)$, we obtain\footnote{Notice that 
$\av{Y_{\bar a}(D_a)}_{p(Y_{\bar a}(D_a)|A=a, Q=q^n, D=d^n)} = \av{Y}_{p(Y|A=a, Q=q^n, D=d^n)}-\textrm{PSE}^{\cancel{a}}_{\bar a a}$. Indeed $\av{Y}_{p(Y|A=a, Q=q^n, D=d^n)}=\theta^y+\theta^y_{a}+\theta^y_{q}q^n+\theta^y_{d}d^n$ and $\textrm{PSE}_{\bar a a}=\theta^y_a$.
This equivalence does not hold in the non-linear setting.}
\begin{align*}
p(Y_{\bar a}(D_a)|A=a, Q=q^n, D=d^n)&=p(Y_{\bar a}(D_a)| Q^*=q^n, D_a=d^n)\\
&=p(Y| A=\bar a, Q=q^n, D=d^n)\,,
\end{align*}
\ie~the counterfactual prediction can be estimated by conditioning $Y$ on $q^n$ and $d^n$, and by replacing $a$ with $\bar a$ in the direct path $A\rightarrow Y$. 

In the case in which $A\rightarrow D$ is also unfair, we have
\begin{align*}
p(Y_{\bar a}(D_{\bar a})|A=a, Q=q^n, D=d^n)&=p(Y_{\bar a}(D_{\bar a})| Q^*=q^n, D_{\bar a}=d^n-\theta^d_aa+\theta^d_a \bar a)\\
&=p(Y| A=\bar a,Q=q^n, D=d^n-\theta^d_aa+\theta^d_a \bar a)\,,
\end{align*}
\ie~the counterfactual prediction can be estimated by conditioning $Y$ on $q^n$ and the corrected version of $d^n$, given by $d^n-\theta^d_aa+\theta^d_a \bar a$. In a more complex scenario in which \eg~$D=f(A, \epsilon _d)$ for a non-linear function $f$ we can sample $\epsilon^{n,m}_d$ from $p(\epsilon_d|a, q^n, d^n)$ and perform a Monte-Carlo approximation of \eqref{eq:counterf}, obtaining
\begin{align*}
p(Y_{\bar a}(D_{\bar a})|A=a, Q=q^n, D=d^n)=\frac{1}{M}\sum_{m=1}^M p(Y| A=\bar a, Q=q^n, D=d^{n,m})\,,
\end{align*}
where $d^{n,m}=f(A=\bar a, \epsilon^{n,m}_d)$ can be interpreted as a corrected version of $d^n$. 

From the discussion above we can deduce that the general procedure for computing the desired counterfactual outcome is to condition $Y$ on the non-descendants of $A$, on the descendants of $A$ that are only fairly influenced by $A$, and on \emph{corrected versions} of the descendants that are (partially) unfairly influenced by $A$. 

\section{Conclusions}
We used causal Bayesian networks to provide a graphical interpretation of unfairness in a dataset as the presence of an unfair causal path. 
We used this viewpoint to revisit the recent debate surrounding the COMPAS pretrial risk assessment tool and, more generally, to point out that fairness evaluation on a model requires careful considerations on the patterns of unfairness underlying the training data. We then showed that causal Bayesian networks provide us with a powerful tool to measure unfairness in a dataset and to design fair models in complex unfairness scenarios.

Our discussion did not cover difficulties in making reasonable assumptions on the structure of the causal Bayesian network underlying a dataset, nor on the estimations of the associated conditional distributions or of other quantities of interest. These are obstacles that need to be carefully considered to avoid improper usage of this framework.

\subsection*{Acknowledgements}
The authors would like to thank Ray Jiang, Christina Heinze-Deml, Tom Stepleton, Tom Everitt, and Shira Mitchell for useful discussions.

\begin{subappendices}
\renewcommand{\thesection}{\Alph{section}}
\vspace{-0.2cm}
\section{Bayesian Networks\label{sec:BN}}
A \emph{graph} is a collection of nodes and links connecting pairs of nodes.
The links may be directed or undirected, giving rise to \emph{directed} or \emph{undirected graphs} respectively.\\[5pt]
A \emph{path} from node $X_i$ to node $X_j$ is a sequence of linked nodes starting
at $X_i$ and ending at $X_j$. A \emph{directed path} is a path whose links are directed and pointing from preceding towards following nodes in the sequence.

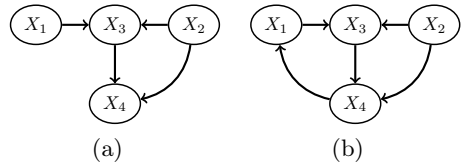
\begin{wrapfigure}[9]{l}{0.5\textwidth}
\vspace{-0.6cm}
\centering
\subfigure[]{\scalebox{0.7}{
\begin{tikzpicture}[dgraph]
\node[ocont] (x2) at (4.5,0) {$X_2$};
\node[ocont] (x1) at (1.5,0) {$X_1$};
\node[ocont] (x3) at (3,0) {$X_3$};
\node[ocont] (x4) at (3,-1.5) {$X_4$};
\draw[line width=1.15pt](x1)--(x3);\draw[line width=1.15pt](x2)--(x3);\draw[line width=1.15pt](x3)--(x4);\draw[line width=1.15pt](x2)to [bend left=35](x4);
\end{tikzpicture}}}
\hskip0.2cm
\subfigure[]{\scalebox{0.7}{
\begin{tikzpicture}[dgraph]
\node[ocont] (x2) at (4.5,0) {$X_2$};
\node[ocont] (x1) at (1.5,0) {$X_1$};
\node[ocont] (x3) at (3,0) {$X_3$};
\node[ocont] (x4) at (3,-1.5) {$X_4$};
\draw[line width=1.15pt](x1)--(x3);\draw[line width=1.15pt](x2)--(x3);\draw[line width=1.15pt](x3)--(x4);\draw[line width=1.15pt](x2)to [bend left=35](x4);\draw[line width=1.15pt](x4)to [bend left=35](x1);
\end{tikzpicture}}}
\vspace{-0.35cm}
\caption{Directed (a) acyclic and (b) cyclic graph.}
\label{fig:GM}
\end{wrapfigure} 

\noindent A \emph{directed acyclic graph} (DAG) is a directed graph with no directed paths starting and ending at the same node. For example, the directed graph in \figref{fig:GM}(a) is acyclic. The addition of a link from $X_4$ to $X_1$ makes the graph cyclic (\figref{fig:GM}(b)). 
A node $X_i$ with a directed link to $X_j$ is called \emph{parent}
of $X_j$. In this case, $X_j$ is called \emph{child} of $X_i$.\\[5pt] 
A node is a \emph{collider} on a path if it has (at least) two parents on that path.
Notice that a node can be a collider on a path and a non-collider on another path. For example, in \figref{fig:GM}(a) $X_3$ is a collider on the path $X_1 \rightarrow X_3 \leftarrow X_2$ and a non-collider on the path $X_2\rightarrow X_3\rightarrow X_4$.\\[5pt]
A node $X_i$ is an \emph{ancestor} of a node $X_j$ if there exists a directed path from $X_i$ to $X_j$. In this case, $X_j$ is a \emph{descendant} of $X_i$.\\[5pt]
A \emph{Bayesian network} is a DAG in which nodes represent random variables and links express statistical relationships between the variables. Each node $X_i$ in the graph is associated with the conditional distribution $p(X_i|\text{pa}(X_i))$, where $\text{pa}(X_i)$ is the set of parents of $X_i$. The joint distribution of all nodes, $p(X_1,\ldots,X_I)$, is given by the product of all conditional distributions, \ie~$p(X_1,\ldots,X_I)=\prod_{i=1}^Ip(X_i|\text{pa}(X_i))$.

In a Bayesian network, the sets of variables ${\cal X}$ and ${\cal Y}$ are statistically independent given ${\cal Z}$ (${\cal X} \ci {\cal Y} \,|\, {\cal Z}$) if all paths
from any element of ${\cal X}$ to any element of ${\cal Y}$ are \emph{closed} (or \emph{blocked}). A path is closed if at least one of the following conditions is satisfied:
\begin{itemize}
\item[(a)] There is a non-collider on the path which belongs to the conditioning set ${\cal Z}$.
\item[(b)] There is a collider on the path such that neither the collider nor any of its descendants belong to the conditioning set ${\cal Z}$.
\end{itemize}

\section{EFPRs/EFNRs and Calibration\label{sec:EFPRFNRC}}
Assume that EFPRs/EFNRs are satisfied, \ie~$p(\hat Y = 1 |A=0, Y=1)=p(\hat Y =1|A=1, Y=1)\equiv p_{\hat Y_1|Y_1}$ and 
$p(\hat Y = 1 |A=0, Y=0)=p(\hat Y =1|A=1, Y=0)\equiv p_{\hat Y_1|Y_0}$. From\\[-14pt]
\begin{align*}
p(Y = 1 |A=0, \hat Y=1)&= \frac{p_{\hat Y_1|Y_1}\overbrace{p(Y=1|A=0)}^{p_{Y_1|A_0}}}{p_{\hat Y_1|Y_1}p_{Y_1|A_0}+p_{\hat Y_1|Y_0}(1-p_{Y_1|A_0})}\,,\\
p(Y = 1 |A=1, \hat Y=1)&=  \frac{p_{\hat Y_1|Y_{1}}p_{Y_1|A_1}}{p_{\hat Y_1|Y_{1}}p_{Y_1|A_1}+p_{\hat Y_1|Y_0}(1-p_{Y_1|A_1})}\,,
\end{align*}
we see that, to also satisfy $p(Y = 1 |A=0, \hat Y=1) = p(Y = 1 |A=1, \hat Y=1)$, we need $\big(\cancel{p_{\hat Y_1|Y_{1}}p_{Y_1|A_1}}+\cancel{p_{\hat Y_1|Y_0}}(1-\cancel{p_{Y_1|A_1}})\big)p_{Y_1|A_0}=\big(\cancel{p_{\hat Y_1|Y_1}p_{Y_1|A_0}}+\cancel{p_{\hat Y_1|Y_0}}(1-\cancel{p_{Y_1|A_0}})\big)p_{Y_1|A_1}$, \ie~$p_{Y_1|A_0} = p_{Y_1|A_1}$. 

\end{subappendices}
\bibliographystyle{plain}
\bibliography{main}

\end{document}